\documentclass[conference]{IEEEtran}
\IEEEoverridecommandlockouts
\usepackage{cite}
\usepackage{amsmath,amssymb,amsfonts}
\usepackage{algorithmic}
\usepackage{graphicx}
\usepackage{textcomp}
\usepackage{xcolor}

\usepackage{tabularx}  
\usepackage{booktabs}  
\usepackage[figuresright]{rotating} 
\usepackage{amssymb}

\usepackage{dsfont}    
\def\BibTeX{{\rm B\kern-.05em{\sc i\kern-.025em b}\kern-.08em
    T\kern-.1667em\lower.7ex\hbox{E}\kern-.125emX}}
\begin{document}

\title{A Shop Floor Production Scheduling Case based on RFID-supported Smart Factory}

\author{
    \IEEEauthorblockN{Zhihui Chen}
    \IEEEauthorblockA{\textit{Department of Mathematics} \\
    \textit{The University of Hong Kong}\\
    Hong Kong, China \\
    zhihuic@connect.hku.hk}
\and
    \IEEEauthorblockN{Yize Sun}
    \IEEEauthorblockA{\textit{Department of Mathematics} \\
    \textit{The University of Hong Kong}\\
    Hong Kong, China \\
    yizesun@connect.hku.hk}
\and
    \IEEEauthorblockN{Yuhao Dong}
    \IEEEauthorblockA{\textit{Department of Mathematics} \\
    \textit{The University of Hong Kong}\\
    Hong Kong, China \\
    dongyh21@connect.hku.hk}
\and
    \IEEEauthorblockN{Zeyu Xiao}
    \IEEEauthorblockA{\textit{Department of Mathematics} \\
    \textit{The University of Hong Kong}\\
    Hong Kong, China \\
    xiaozeyu@connect.hku.hk}
\and
    \IEEEauthorblockN{Ray Y. Zhong}
    \IEEEauthorblockA{\textit{Department of Industrial and Manufacturing Systems Engineering} \\
    \textit{The University of Hong Kong}\\
    Hong Kong, China \\
    zhongzry@hku.hk}
}

\maketitle

\begin{abstract}
Radio frequency identification (RFID) technology has been widely implemented for real-time data collection in manufacturing shop floors, which, in turn, can be used to support dynamic shop floor production planning and scheduling. Within such an environment, uncertainty in operation and production processes collectively contribute to the dynamicity in manufacturing, thereby hampering the scheduling system from achieving maximal utility. To highlight the importance of handling such uncertainty, this paper addresses the problem of dynamic shop floor scheduling for a real-life case smart factory equipped with RFID technology. Feasible production sequence mining and real-time processing rate estimation are conducted on RFID-collected production data to quantify the operation and production uncertainties. A deep reinforcement learning approach based on the RFID data analysis is then presented for shop floor production scheduling. Simulation studies based on real-life case data have demonstrated the feasibility and practicality of the proposed dynamic production scheduling framework. Specifically, it is observed that the proposed framework outperforms existing dispatch methods in terms of minimizing operation makespan, including first in first out (FIFO), last in first out (LIFO) and deep Q network (DQN).
\end{abstract}



\section{Introduction}
In shop floor manufacturing, non-processing process (idle and waiting times) cause waste in resources and potential delays in product delivery. Shop floor production scheduling therefore plays a pivotal role in smart manufacturing by reducing total production time cost and improving operation efficiency \cite{rfid_schedule}. During manufacturing, the shop floor material flow is scheduled through assigning production jobs/raw materials with individualized processing sequences to minimize the time cost for optimal production efficiency.


The complexity of production scheduling mainly lies in the combination of dynamic production process, and the real-time shop floor visibility \cite{ray_dynam_schedule,real_time_vis_schedule}.  
Existing literature usually places two types of preconditions on the scheduling case due to limitations in scheduling method and the availability of real-time data. Firstly, the processing sequence for jobs going through each machine is predefined when formulating the scheduling algorithm for mathematical simplicity. For job shop scheduling research \cite{job_shop_review}, the operation sequence  is totally fixed. For open shop scheduling, the operation sequence for each job is arranged as completely random \cite{open_shop_review}. Secondly, the processing time of jobs on machines are predetermined. Due to the lack of real-time visibility of production shop floor, most research sets a pre-determined processing time for each job on each machine for the scheduling case \cite{real_time_vis_schedule,cirp_scheduling}. These hypothetical assumptions in the current research cannot fully describe the actual shop floor production environment where complex and dynamic manufacturing state often occur. Firstly, pre-determined job processing sequence are not the common case in medium-small size manufacturers. Research has shown that, alternative routing for production jobs can be common in real-world manufacturing shop floors \cite{alternative_root}.  Secondly, pre-determined job processing time does not reflect the dynamic environment in real-life shop floor where operation speed can affected by human and non-human centric factors \cite{chen_production_efficiency}.

In recent years, Radio frequency identification (RFID) em-erges as a new form of real-time production data collection technology, enabling real-time tracking and monitoring of production processes \cite{zhong2013real, zhong2013rfid}. With smart RFID tags attached to processed materials, RFID readers are able to automatically collect real-time data for product tracking and identification. With its advantage in speed and automation, RFID positions itself as an advanced approach for attaining real-time visibility on the shop floor \cite{a21,zhong2013rfid_planing}. A RFID-supported smart factory, therefore, possesses the ability to monitor the real-time shop floor status and accumulate manufacturing big data, enabling a more informed shop floor production scheduling practice.

In order to bridge the gap between the constrained solutions of existing methods and real-world applications, this paper proposes an RFID data analysis and reinforcement learning (RL)-supported shop floor scheduling framework. RFID data analysis is conducted to provide realistic information for easing the above preconditions. Firstly, feasible production sequence mining is applied on the RFID-collected shop floor production data to address the interchangeability of production processes. The flexibility to consider alternative routing provides the potential to improve productivity based on the unique circumstances of each production cycle. Secondly, targeting the pre-determined job processing time, a real-time production rate estimation is provided based on RFID data analysis. It supports a more informed scheduling decision by providing a real-time estimation of current machine processing time for certain type of jobs. Given the RFID data analysis results, a deep Q network (DQN)-based RL model is proposed to conduct dynamic shop floor scheduling. The DQN is informed with real-time RFID-collected shop floor status and RFID data analysis results for conducting scheduling decisions on shop floor jobs.

The rest of the paper is structured as follows. Section 2 presents the proposed scheduling framework and RL environment. Section 3 reports a real-life numerical study on a RFID-support manufacturing company's case. The last section summarises the findings and gives a research outlook.

\section{Methodology}

\subsection{RFID data analysis for dynamic shop floor scheduling} \label{sec:analysis}

\begin{figure*}[htbp]
    \centering
    \includegraphics[width=0.7\linewidth]{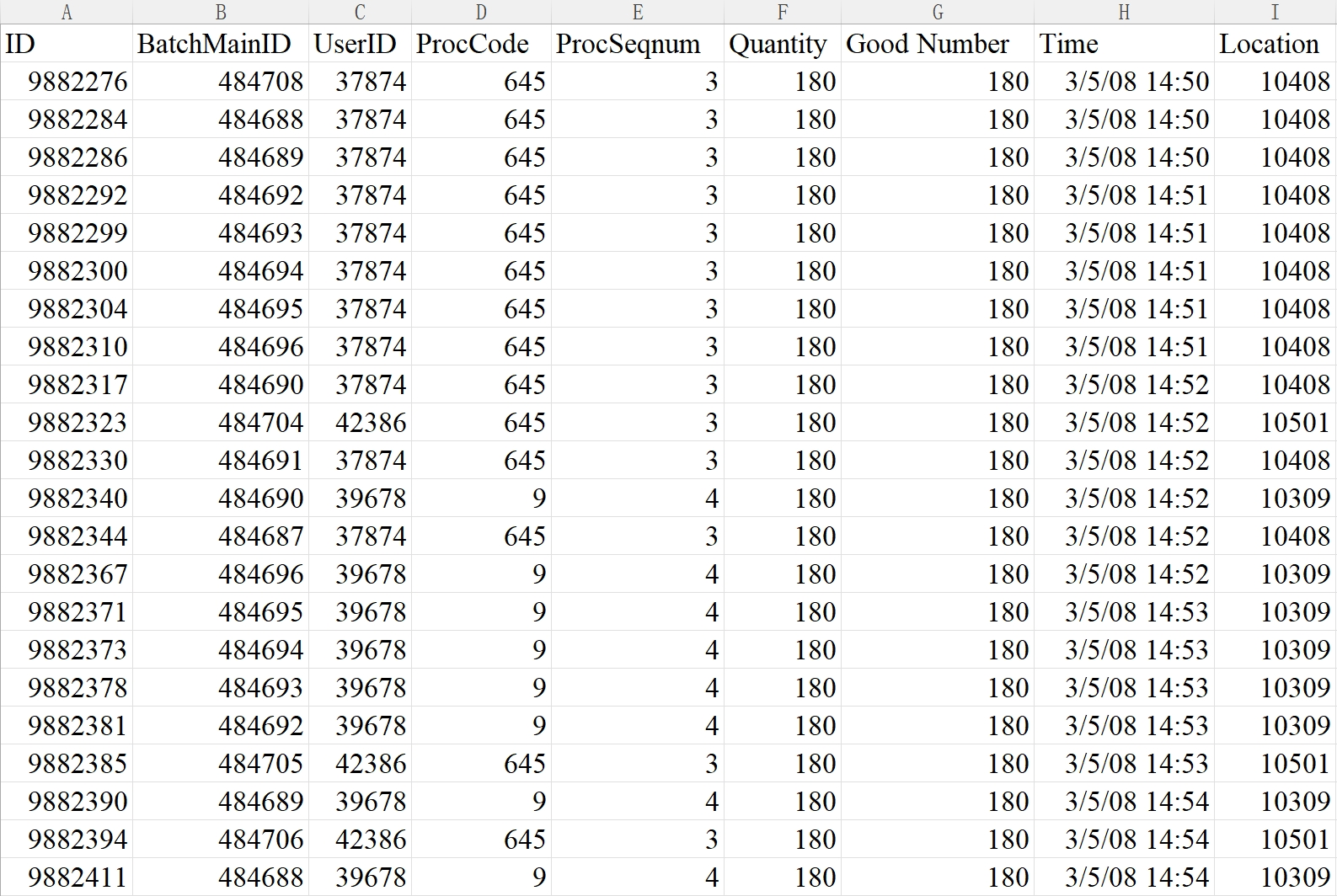}
    \caption{Sample Data}
    \label{fig:raw_demo}
\end{figure*}

\subsubsection{Data overview}
This study is based on an RFID-collected manufacturing dataset containing 413,472 data records. The data were collected from Huaiji Dengyun Auto Parts (Holding) Co., Ltd., a manufacturing company primarily engaged in the research and development, production and sales of automobile engine valves \cite{rfid_dataset}. The company applied RFID technology in their assembly line to record relevant data reflecting the whole shop floor production process for production management improvement. Partially shown in Fig. \ref{fig:raw_demo}, the shop floor production dataset contains records range from March to July 2008 and include the following key attributes:

\begin{enumerate}
    \item ID: Representing the Auto-generated ID in SQL database;
    \item BatchMainID: Representing the ID of a batch of products;
    \item UserID: Representing the ID number of a specific operator;
    \item ProcCode: Representing the code for a particular production process, such as milling or drilling;
    \item ProcSeqnum: Representing the sequence of the production processes;
    \item Quantity: Representing the total quantity of a batch of products, with values ranging from 0 to 180.
    \item Good Number: Representing the number of products that passed quality inspection, with values ranging  from 0 to 180;
    \item Time: Representing a timestamp reflecting the finish time of a
    process (e.g. '2008-03-05 13:12:00');
    \item Location: Representing a particular machine.
\end{enumerate}

\subsubsection{Feasible production sequence mining}

In dynamic shop floor scheduling, jobs must pass through each machine in a feasible production sequence. By analyzing the RFID data, it can be found that the machining order of different ProcCodes under the same BatchMainID may be different. Table \ref{tab:proccode} presents a few successful production records from the RFID dataset. It demonstrates the interchangeability of production sequence for the same production task. This is due to the real-life manufacturing case where workers may switch the order of machining, such as milling and drilling, during the process of producing a workpiece. It can be found that, interchanging specific production processes does not affect the final production quality. Shop floor scheduling, therefore, can be optimized based on the alternative feasible production sequence when facing production bottlenecks on certain stage of prodcution.

\begin{table}
  \centering
  \caption{Different ProcCode Sequence}
  \label{tab:proccode}
  \begin{tabular}{ccccccr}
    \toprule
    BatchMainID & \multicolumn{5}{c}{ProcCode}\\
    \midrule
    484587 & 2 & 645 & 9 & 705 & 213\\
    495932 & 2 & 9  & 705   & 213 & 645\\
    495933 & 2 & 9   & 645 & 705 & 213 \\
    495934 & 2 & 645 & 705 & 213 & 9 \\
    \bottomrule
  \end{tabular}
\end{table}

\subsubsection{Production rate regression analysis}
In the realm of shop floor scheduling, knowing the machining duration of jobs on machines in advance proves challenging. Nevertheless, a viable approach involves estimating these time frames through regression analysis using real-time RFID-collected data. Examination of this RFID data reveals significant variations in the machining time of identical ProcCodes under the same BatchMainID. This variation arises from factors such as rest periods and shift changes, which impact the recording of processing times during workpiece production. To refine and structure the data for further analysis, a comprehensive data processing step is conducted.

This data processing included the introduction of new attributes derived from the existing data attributes. These attributes contains diverse elements, such as segmenting time periods (morning, noon, afternoon, night, midnight) based on timestamp information, identifying holidays and weekends using date information, and estimating production rate metrics through predefined equations. 



\begin{enumerate}
    \item Date: The corresponding date of the Time attribute.
    \item Holiday: Whether the production date is on a Chinese official holiday.
    \item Weekend: Whether the production date is on weekend.
    \item Time Slot: Representing the corresponding daily time slot of production time. Morning: 6 - 10am; Noon: 11am - 1pm; Afternoon: 2 - 6pm; Night: 7 - 11pm; Midnight: 0 - 5am;
    \item Time Cost:  Representing the total time consumed to produce certain number of products.
    \item Production Rate: The average speed of production within a specific hourly time interval, as shown in (\ref{eq:rate}).
\end{enumerate}

After analyzing the raw data, the subsequent step involves the implementation of a regression analysis. The primary objectives of regression analysis encompass delineating relationships between variables, managing predictor variables for a given response value, and predicting responses based on these predictors.

\begin{equation}
    \mathrm{Produciton \: rate} = \frac{\sum_t\mathrm{Quantity}}{\sum_t\mathrm{Time \: Cost}},  \;
    \mathrm{where \:} t \in 1h
    \label{eq:rate}
\end{equation}

In this context, linear regression functions as a statistical tool to establish connections between various attributes and the average production rate. Through linear regression, we can estimate the coefficients of independent variables (attributes) and assess their significance in elucidating variations in the average production rate. To execute this analysis, historical RFID-generated data and enriched attributes such as Date, Holiday, Weekend, Time Slot, Time Cost, and Average Production Rate will be utilized. The average production rate serves as the dependent variable, while the other attributes function as independent variables.

The analysis centers on evaluating the average production rate for a machine during specific types of operations (ProcCode) within designated time slots of 60 minutes length. Employing a lasso regression approach aids in determining the significance of inputs. This method offers insights into the sequential entry of inputs into the regression model. The weights assigned to each attribute in the Lasso regression are indicative of their quantitative impact on production rate. Through the utilization of RFID data in feasible production sequence mining and production rate regression analysis, valuable insights concerning production sequences and job processing times on the machine can be concluded for addressing challenges in shop floor manufacturing scheduling.

\begin{figure*}[!h]
    \centering
    \includegraphics[width=\linewidth]{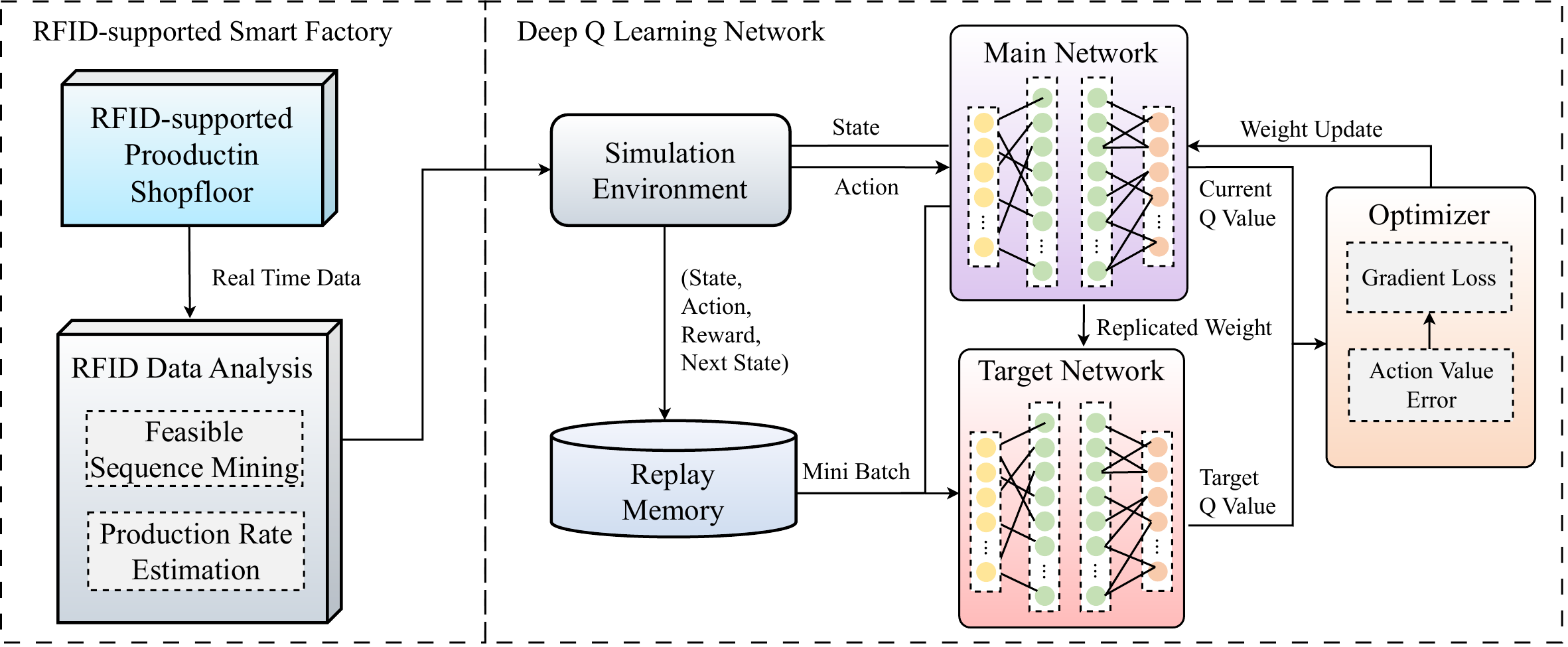}
    \caption{DRL-based dynamic shop floor scheduling framework}
    \label{fig:dqn}
\end{figure*}

\subsection{DRL-based dynamic shop floor scheduling}
Fig. \ref{fig:dqn} demonstrates the overall framework of the DRL-based dynamic shop floor scheduling approach. With dynamicity in production sequence and batch production rate quantified by RFID data analysis, a decision model capable of handling such flexibility is needed in addressing the dynamic shop floor scheduling problem. Possessing the capability of adapt and learn from dynamic environment and making data-informed decisions, deep reinforcement learning (DRL) presents itself as a promising approach to address this challenge. Here, a DRL-based shop floor scheduling framework is proposed: to learn a shop floor scheduling policy based on shop floor simulation environment and real-life manufacturing data given an objective and then conduct shop floor scheduling by allocating jobs to machines based on the learned policy.


In the shop floor scheduling task, there exist jobs $J$ and machines $M$ where each machine represent a stage of production. Each job $J_{i} \in J$ must go through each machine in $M$ in a feasible production sequence mined by RFID data analysis. The processing time for any job at any machine is deterministic but not foreseeable by the decision making agent in advance. Through the proposed production rate sensitivity analysis, processing time can be estimated to provide additional information for the decision making. All machines can only operate on a given job at a time and the operation can not be interrupted once started. The task is considered completed when all jobs are finished and the total time cost is denoted as the makespan of production: $MS$. 

The DRL-based model aims to perform scheduling by learning a optimal policy from its interaction with the environment. Specifically, the deep Q learning is leveraged to identify an optimal action-selection policy for the shop floor scheduling task. Key components are defined to develop the DRL-based framework.
\subsubsection{States}
The optimal scheduling action is determined based on the observation of the current state $s \in S$. In the proposed framework, the current progress of production is represented by a ($J + 2|M|$) matrix, containing $3 + M$ attributes for each row (i.e. every job): (1) a Boolean attribute o represent if the job has been allocated; (2) the estimated remaining time on the current machine operation gained from sensitivity analysis; (3) the true remaining time on the current machine operation based on historical data; this attribute is used for environment simulation and will be masked for deep Q network to ensure the decision agent can not foreseen the finishing time; The $3:(3+M)$ attributes representing the production progress of the job on all machines in Boolean form; 
\subsubsection{Actions}
The action is defined as the act of allocation of one job $J_{i} \in J$ to machine $m_{i} \in M$ ($a \in  A_{allocate}$) or the act of waiting for production ($a \in A_{wait}$). The representation of the whole action space is a ($|J| \times |M|$ + 1) vector for each possible action. Each job must go through all machines in a feasible order mined from RFID-collected production data and cannot be allocated when the job is currently in process or the target machine is occupied. This constraint is realized by a mask posted on the Q value table to stop decision making agent in deep Q network from choosing infeasible actions.
\subsubsection{Rewards}
The reward $R(s, a, s') \in R$ represents the immediate feedback for the decision maker received after transitioning from $s$ to $s'$ due to action $a$. With In dynamic shop floor scheduling, the reward function is formulated in (\ref{eq:reward}). $MS_{s}$ represents the current makespan at state $s$, $\mathds{1}_{A}(a)$ is the indicator of $a \in A_{allocate}$ and $\mathds{1}_{S}(s')$ is the indicator of $s'$ being the end state of production.
\begin{equation}
\centering
    R(s, a, s') = - (MS_{s'}-MS_{s})  - 0.5 \times \mathds{1}_{A}(a) + 1000 \times \mathds{1}_{S}(s') 
    \label{eq:reward}
\end{equation}

Given the above DRL settings, the deep Q network performs reinforcement learning to search for an optimal scheduling policy through updating the Q value according to (\ref{eq:q_learning}).
\begin{equation}
Q(s, a) =  Q(s, a) + \alpha * [R(s, a, s') + \gamma * \max_{a'} Q(s', a') -  Q(s, a) ] 
\label{eq:q_learning}
\end{equation}
where $\alpha \in [0,1]$ denotes a learning rate representing how fast agents learn from new knowledge and override old knowledge. During the learning phase, the epsilon-greedy algorithm ($\epsilon$ in the range of [0, 1]) is employed to balance exploration and exploitation. With a probability $\epsilon$, the agent exploits its current knowledge of action values, while with a probability $1-\epsilon$, the agent explores non-greedy actions to enhance its action value estimations. Artificial neural networks are utilized to estimate action values as function approximators without explicit state space design. Network parameters are updated using a loss function based on replay memory through gradient descent steps. The loss function is defined as $(y_m - Q(s_m, a_m))^2$, where $y_m$ is equal to $r_m$ if the state terminates at time $m+1$ and is calculated differently if the state does not terminate at that time. The target network's parameters, denoted as $\hat{\theta}$, are updated periodically at fixed intervals.

 \section{Numerical Studies}
In this study, the proposed shop floor scheduling framework is evaluated on real life scheduling case of Huaiji manufacturing company. The RFID-supported smart factory in Huaiji produced massive RFID-collected production data containing multiple types of product at different shop floors. For simplicity, this study focuses on the production of type I product which requires 5 different production process, denoted by Production Code 2, 645, 9, 705, and 213. Specifically, a 5 machine 20 job scheduling environment is considered where each machine conducts a production process for the type I product and the number of unprocessed jobs is 20. Among the dynamic shop floor scheduling framework, RFID production data analysis is firstly conducted to mine the feasible production sequence and analysis the influence factors for production rate of type I product within each 5 process. Supported with the above information, the proposed DRL-based scheduling model is tested on virtual production environment simulated by historical manufacturing data. The numerical experiments are conducted on a Lenovo Y9000X laptop with Intel Core i7-12700H CPU, NVIDIA RTX30060 GPU and 40 GB of RAM. The programming was completed in Python and CUDA-supported Pytorch.

\subsection{RFID data analysis: Feasible production sequence mining}
Feasible production sequence mining aims to discover feasible production sequence of the required processes in real-life shop floor manufacturing. With the focus on type I product, the production process includes 5 different production process denoted by Production Code 2, 645, 9, 705, and 213 in RFID-collected manufacturing dataset. By tracking the production sequence each batch of product, the RFID-collected data reveals the feasible sequences in producing type I product, which are concluded and shown in Table \ref{table:sequence}.

\begin{table}[h]
    \centering
    \caption{Type I product feasible production sequence}
    \begin{tabular}{ccccc|c}
        \toprule
        \multicolumn{5}{c}{Production Sequence} & Batch Count \\
        \midrule
        2.0 & 645.0 & 9.0 & 705.0 & 213.0 & 41 \\
        2.0 & 645.0 & 9.0 & 213.0 & 705.0 & 3 \\
        645.0 & 9.0 & 2.0 & 705.0 & 213.0 & 13 \\
        645.0 & 9.0 & 2.0 & 213.0 & 705.0 & 1 \\
        9.0 & 2.0 & 645.0 & 705.0 & 213.0 & 8 \\
        2.0 & 9.0 & 645.0 & 705.0 & 213.0 & 4 \\
        \bottomrule
    \end{tabular}
    \label{table:sequence}
\end{table}
A clear pattern regarding  the production sequence can be observed: the first three processes and the last two processes can be interchanged internally without hampering the real life production. In addition, a constraint can be observed: the first three processes must be done before conducting the last two. Based on the data mining result, we can concluded that in real life manufacturing case, the production of type I product does not necessarily require a fixed production sequence, on the contrary, certain degree of interchangeability in production sequence exists in real life shop floor manufacturing. Such flexibility in production sequence has a great potential in contributing to more efficient shop floor production scheduling by improving machine utilization and avoid production bottleneck.

\subsection{RFID data analysis: Production rate regression analysis}
Regression analysis is employed to examine the impact of various manufacturing factors on production rate. Accurately estimating machine production speed is crucial for real-time shop floor scheduling in dynamic production environments, especially when the true finish time of a product cannot be predicted in advance. This study focuses on analyzing the sensitivity of production processes 2, 645, 9, 705, and 213 during the manufacturing of type I products. As outlined in Section \ref{sec:analysis}, potential influencing factors such as date and time slot of work, production location, and batch quantity are derived from the raw RFID-collected manufacturing data. The RFID production rate for type I products consists of 400 records. Training set contains randomly selected 70\% of the whole records while test set contains the rest 30\% of records. Lasso regression is conducted to provide estimation on the machine production rate for each process. The regression coefficients and its P-value in F test are shown in Table \ref{table:lasso_results}.

\begin{table}[h]
\caption{Lasso Regression Results}
\begin{tabular*}{\hsize}{@{\extracolsep{\fill}}ccccl@{}}
\toprule
Variable & Coefficient & Standard Error & P-value \\
\midrule
Constant & 9.4638 & 0.861 & 0.000 \\
Date & -0.0323 & 0.020 & 0.101 \\
ProcCode & 0.0196 & 0.005 & 0.000 \\
Morning & -0.1980 & 0.821 & 0.000 \\
Noon & -0.5257 & 1.071 & 0.002 \\
Afternoon & -1.5887 & 0.682 & 0.000 \\
Night & 0.5257 & 0.838 & 0.000 \\
Midnight & -1.4579 & 0.692 & 0.000 \\
Location & -0.6947 & 0.223 & 0.002 \\
Weekend & -0.5066 & 0.077 & 0.000 \\
Holiday & -0.1200 & 0.258 & 0.465 \\
Quantity & 0.0075 & 0.000 & 0.000 \\
\bottomrule
\end{tabular*}
\label{table:lasso_results}
\end{table}


\begin{table}[h]
    \centering
    \caption{Performance Comparison of different scheduling methods}
    \begin{tabularx}{\columnwidth}{XXX}
        \toprule
        Method & Sum reward & Makespan (minute) \\
        \midrule
        Proposed & 574.5 & \textbf{476} \\
        DQN & 577.5 & 485 \\
        FIFO & 567.5 & 495  \\
        LIFO & 578.5 & 484 \\
        Random & 567.5 & 495 \\
        \bottomrule
    \end{tabularx}
    \label{table:ressult}
\end{table}

\begin{figure*}[h]
    \centering
    \includegraphics[width=0.8\linewidth]{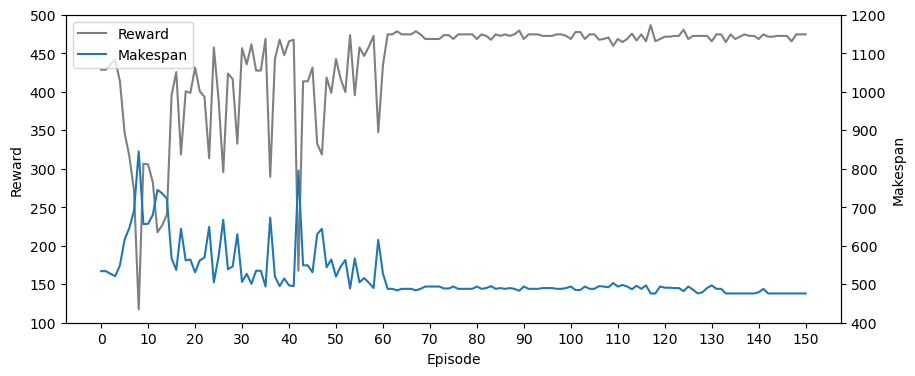}
    \caption{Deep Q network model training curve}
    \label{fig:train}
\end{figure*}

As shown in Table \ref{table:lasso_results}, "ProcCode," temporal factors ("Morning," "Noon," "Afternoon," "Night," "Midnight"), and "Quantity" exhibit significant in T test (P-value < 0.05), implying a certain influence while date related attributes ("Date", "Holiday") demonstrate uncertain influence on the machine production rate. The Lasso regression achieves a mean absolute percentage error (MAPE) of 40.23\% and a $R^{2}$ score of 0.80 on the test set with 120 records. 



\subsection{DRL-based production scheduling}
The proposed DQN model is provided with production data analysis results including the estimated job processing time and current alternative production sequence to better discover efficient scheduling solutions for minimizing the total makespan of shop floor production. Scheduling experiments are conducted in simulation environment supported by real-life shop floor data. The scheduling experiment is set as a shop floor production case with 15 type I jobs and  5 machines each represents a type of necessary process. 

Fig. \ref{fig:train} demonstrates the training process of the deep Q network as part of the scheduling framework. It is shown that the network converges after around 140 epochs of training with reward and production makespan reaches a stable level. It is shown that the training reward converges to a optimal level in around 80 epochs, same as the production makespan. As shown in Table \ref{table:ressult}, the proposed DRL-based dynamic shop floor scheduling framework outperforms existing dispatch methods in terms of minimizing operation makespan, including deep Q learning network (DQN), first in first out (FIFO), last in first out (LIFO) and random allocation. Is is shown that, the proposed model outperforms other deep learning-based and classical scheduling approaches.

\section{Conclusions}
In conclusion, this paper contributes to the field of smart manufacturing by addressing the uncertainties in shop floor scheduling through RFID technology. The feasibility of production sequence mining and the impact of dynamic processing times on production rates are explored. The proposed DRL-based dynamic shop floor scheduling framework is demonstrated to outperform existing dispatch methods, including FIFO, LIFO, DQN and random allocation. The model's ability to adapt to real-time data and optimize production sequences underscores its potential for enhancing efficiency and minimizing makespan in dynamic manufacturing environments. Future research in this area should focus on further bridging the gap between theoretical models and real-world production scenarios, incorporating adaptability and reflecting the complexities of practical manufacturing operations.

\bibliographystyle{unsrt}
\bibliography{references}

\begin{thebibliography}{10}

\bibitem{rfid_schedule}
Xiaoqiang Wu, Songling Tian, and Lei Zhang.
\newblock The internet of things enabled shop floor scheduling and process
  control method based on petri nets.
\newblock {\em IEEE access}, 7:27432--27442, 2019.

\bibitem{ray_dynam_schedule}
Yingfeng Zhang, George~Q Huang, Shudong Sun, and Teng Yang.
\newblock Multi-agent based real-time production scheduling method for radio
  frequency identification enabled ubiquitous shopfloor environment.
\newblock {\em Computers \& Industrial Engineering}, 76:89--97, 2014.

\bibitem{real_time_vis_schedule}
T~Wang, YF~Zhang, and DX~Zang.
\newblock Real-time visibility and traceability framework for discrete
  manufacturing shopfloor.
\newblock In {\em Proceedings of the 22nd International Conference on
  Industrial Engineering and Engineering Management 2015: Core Theory and
  Applications of Industrial Engineering (Volume 1)}, pages 763--772. Springer,
  2016.

\bibitem{job_shop_review}
Anant~Singh Jain and Sheik Meeran.
\newblock Deterministic job-shop scheduling: Past, present and future.
\newblock {\em European journal of operational research}, 113(2):390--434,
  1999.

\bibitem{open_shop_review}
Ellur Anand, Ramasamy Panneerselvam, et~al.
\newblock Literature review of open shop scheduling problems.
\newblock {\em Intelligent Information Management}, 7(01):33, 2015.

\bibitem{cirp_scheduling}
Ray~Y Zhong, George~Q Huang, QY~Dai, T~Zhang, TY~Luo, and DP~Lin.
\newblock A real-time planning and scheduling model in rfid-enabled
  manufacturing.
\newblock In {\em 46th CIRP conference on manufacturing systems}, pages 29--30,
  2013.

\bibitem{alternative_root}
TN~Wong, CW~Leung, Kai-Ling Mak, and Richard~YK Fung.
\newblock Dynamic shopfloor scheduling in multi-agent manufacturing systems.
\newblock {\em Expert Systems with Applications}, 31(3):486--494, 2006.

\bibitem{chen_production_efficiency}
Chen Zhihui, Xiao Zeyu, Sun Yize, and Zhong Ray~Y.
\newblock Production efficiency analysis based on the rfid-collected
  manufacturing big data.
\newblock {\em Manufacturing Letters}, in press.

\bibitem{zhong2013real}
Ray~Y Zhong, George~Q Huang, QY~Dai, T~Zhang, TY~Luo, and DP~Lin.
\newblock A real-time planning and scheduling model in rfid-enabled
  manufacturing.
\newblock In {\em 46th CIRP conference on manufacturing systems}, pages 29--30,
  2013.

\bibitem{zhong2013rfid}
Ray~Y Zhong, Z~Li, LY~Pang, Y~Pan, Ting Qu, and George~Q Huang.
\newblock Rfid-enabled real-time advanced planning and scheduling shell for
  production decision making.
\newblock {\em International Journal of Computer Integrated Manufacturing},
  26(7):649--662, 2013.

\bibitem{a21}
Qingyun Dai, Runyang Zhong, George~Q. Huang, Ting Qu, T.~Zhang, and T.~Y. Luo.
\newblock Radio frequency identification-enabled real-time manufacturing
  execution system: a case study in an automotive part manufacturer.
\newblock {\em International Journal of Computer Integrated Manufacturing},
  25(1):51--65, 2012.

\bibitem{zhong2013rfid_planing}
Ray~Y Zhong, Z~Li, LY~Pang, Y~Pan, Ting Qu, and George~Q Huang.
\newblock Rfid-enabled real-time advanced planning and scheduling shell for
  production decision making.
\newblock {\em International Journal of Computer Integrated Manufacturing},
  26(7):649--662, 2013.

\bibitem{rfid_dataset}
Qingyun Dai, Runyang Zhong, George~Q Huang, Ting Qu, Ting Zhang, and TY~Luo.
\newblock Radio frequency identification-enabled real-time manufacturing
  execution system: a case study in an automotive part manufacturer.
\newblock {\em International Journal of Computer Integrated Manufacturing},
  25(1):51--65, 2012.

\end{thebibliography}

\end{document}